\documentclass[11pt]{article}
\usepackage{coling2020}
\usepackage{times}
\usepackage{url}
\usepackage{latexsym}
\usepackage{amssymb}
\usepackage{graphicx}
\usepackage{multirow}
\usepackage{amsmath}
\usepackage{pifont}
\newcommand{\cmark}{\ding{51}}%
\newcommand{\xmark}{\ding{55}}%

\usepackage{xcolor}
\newcommand{\corrval}[2]{{#2}}

\usepackage{courier}
\usepackage{url}
\usepackage{hyperref}

\title{Improving Sentiment Analysis over non-English Tweets using Multilingual Transformers and Automatic Translation for Data-Augmentation}

\author{
	Valentin Barriere \\
	European Commission – DG-JRC \\
	Via Enrico Fermi, 2749 \\
	21027 Ispra (VA), Italy \\
	{\tt name.surname@ec.europa.eu} \\\And
	Alexandra Balahur \\
	European Commission – DG-JRC \\
	Via Enrico Fermi, 2749 \\
	21027 Ispra (VA), Italy \\
	{\tt name.surname@ec.europa.eu} \\}

\date{}

\begin{document}
	\maketitle
	\begin{abstract}
		Tweets are 
		specific text data when compared to general text. Although sentiment analysis over tweets has become very popular in the last decade for English, it is still difficult to find huge annotated corpora for non-English languages. The recent rise of the transformer models in Natural Language Processing allows to achieve unparalleled performances in many tasks, but these models need a consequent quantity of text to adapt to the tweet domain. We propose the use of a multilingual transformer model, that we pre-train over English tweets and apply data-augmentation using automatic translation to adapt the model to non-English languages. Our experiments in French, Spanish, German and Italian suggest that the proposed technique is an efficient way to improve the results of the transformers over small corpora of tweets in a non-English language. 
	\end{abstract}
	
	\section{Introduction}
	\label{sec:intro}
	
	
	Monitoring social media at the scale of a continent, like Europe, requires to process multiple languages. Data from Twitter is noisy and can drastically change in term of words distribution in one language when compared with general texts. It becomes even more challenging when trying to tackle a multilingual task.
	
	In accordance with the RGPD, it is impossible to make available tweets that have been deleted by their authors. This makes it more difficult to find tweet corpora annotated in sentiment. The SemEval challenges assure a big database of tweets in English, with a total of more than 62k examples annotated at the level of the tweet \cite{Rosenthal2018}. For other languages, \corrval{it becomes}{this is} more complicated. 
	
	Bidirectional transformers like BERT \cite{Devlin2018} revolutionized the world of Natural Language Processing (NLP). Even when pre-trained over text from general domain, these models need a substantial amount of data to \corrval{be trained, which can be problematic}{adapt to a domain where the syntax is different, like Twitter}. 
	
	This paper presents the experiments carried out on several datasets of tweets in five different languages: English, French, Spanish, German and Italian. The general idea is pretty simple: instead of using a monolingual model, we chose to use a multilingual model that we can train over a large dataset of English tweets, over the original non-English tweets and over their automatic translations.  
	We chose the multilingual transformer model XLM-RoBERTa from \cite{Lample2019} with a data-augmentation technique using machine translation. 
	We investigated the effects of pre-training with English data and data-augmentation. We also compared performances of multilingual models against their monolingual French \cite{Martin2020} and English \cite{Liu2019} counterparts and found interesting improvements. 
	
	%
	%
	\blfootnote{
		%
		%
		\hspace{-0.65cm}  
		Place licence statement here for the camera-ready version.
		%
		%
		%
		%
	}

	\section{State of the Art}
	\label{sec:sota}
	
	The related works section of this paper is shared between  multilingual sentiment analysis, data-augmentation and sentiment analysis over tweets.

	
	We count several challenges tackling sentiment analysis over tweets like SemEval \cite{Nakov2013} in English, TASS \cite{Villena-Roman2013} in Spanish or DEFT \cite{Hamon2015} in French. 
	For the last years, the neural networks are ruling the sentiment analysis over tweets. \cite{Cliche2017} won the sentiment polarity subtask of the SemEval-2017 challenge \cite{Rosenthal2018} using a neural network approach. 
	\cite{Singh2019} improve the state-of-the-art with a different incorporation of the emojis: they use the descriptions instead of their unicodes before a BLSTM with word embeddings. 
	Finally, \cite{Nguyen2020} proposed a BERT model with the RoBERTa pre-training procedure \cite{Liu2019} over 850M English tweets. This model gives state-of-the-art results on the SemEval-2017 dataset, but like all the other models, it is not adapted to multilingual data. 
	
	Although data-augmentation is not as developed for textual data as it is for images, 
	there are ways to apply it to text \cite{Wei2020}. 
	\cite{Sennrich2016} augment their datasets using back-translation for pairing sentences for a MT task. \cite{Kobayashi2018} uses language models in context in order to create new plausible sentences. \cite{Fadaee2017} use data-augmentation for MT, by automatically translating at the level of words using plausible substitutions in order to create a new sentence.
	
	
	
	The work that is closest to the one we are presenting in this article remains that of \cite{Balahur2013,Balahur2014} who 
	tackle a sentiment analysis task over multilingual tweets. They show that use of  multilingual, machine-translated data can help to better distinguish relevant features for sentiment classification, using SVM models with Bag-of-N-Grams. We distinguish from this work by using real datasets for testing instead of artificially created test sets made of translated tweets re-edited by humans. 
	

	\section{Proposed Method} \vspace*{-.2cm}
	\label{sec:method}
	
	The method we propose is very simple. It basically consists in using a multilingual model instead of a monolingual model, pre-trained it over available annotated English tweets, and combine that with a data-augmentation technique that uses automatically translated tweets. With this augmentation, we have each tweet in five examples, in five different languages. The languages that we use for the tweets are the same languages that the datasets we used to test the tweets: French, English, German, Spanish and Italian. 
	
	\subsection{Pre-training over External Datasets} \vspace*{-.1cm}
	
	As we said earlier, we found it was more difficult to find tweet datasets in languages that were not English. We then investigated the potential of using multilingual model pre-trained over English tweets only, and over English tweets automatically translated in other languages. 
	
	We investigated the effect of using other available English datasets with multilingual model. To that end, we pre-trained the neural network with tweets annotated for the SemEval-2013 to SemEval-2016 challenges. We used the original tweets in English, but also their automatic translations in the 4 other languages we studied. 
	This leads to a total of 47762 tweets, and 238 810 using data-augmentation with automatic translation. The 2000 tweets from the devtest of SemEval-2016 were used as development set and the test set from SemEval-2017 was used as test set. 

	\subsection{Data Augmentation and Multilingual Training} \vspace*{-.1cm}
	
	Translating the tweets into other languages allows our model to see 5 times the amount of data that it should have originally seen. 
	The translations from the source language to the 4 other languages were made by the automatic translation tool of the \href{https://ec.europa.eu/info/departments/translation_en}{European Commission}, which is comparable to Google Translate. You can find examples of tweets and their translation in Table \ref{tab:translated_examples}. It is important to note that the quality of the translation is not optimal since the translation has been learned over general text and tweets can be noisy data containing abbreviations, and modernisms. 
	
	\begin{table}[]
		\resizebox{\textwidth}{!}{%
			\begin{tabular}{l|l}
				Lang.               & Tweet \\ \hline \hline
				\textbf{English} &     I'd rather dump gasoline all over myself and run into a burning building than use Excel.  \\
				French             &    Je préférerais jeter de l’essence partout et tomber dans un immeuble en feu plutôt que d’utiliser Excel.   \\
				German             &   Ich würde lieber Benzin auf mich werfen und in ein brennendes Gebäude laufen, als Excel zu benutzen.    \\
				Spanish            &    Prefiero tirar gasolina sobre mí mismo y correr hacia un edificio en llamas que usar Excel.   \\
				Italian            &    Preferirei buttarmi la benzina addosso e correre in un edificio in fiamme piuttosto che usare Excel.  \\ 
			\end{tabular}%
		}
		\caption{Examples of automatically translated tweets (original language in bold)}
		\label{tab:translated_examples}
	\end{table}
	
	
	Finally, we always fine-tune the model over the non-English target datasets. The results of the models that are not fine-tuned on the target languages are poor and not even reported in this article. 
	

	\section{Experiments and Results} \vspace*{-.1cm}
	\label{sec:expe}
	
	\subsection{Methodology}
	
	The pre-trained models that we used were made available online using the \texttt{transformers} library \cite{Wolf2019}. The same library has been employed for the training of the models. 
	We used the Adam algorithm \cite{Kingma2014} with early stopping for the optimization of the training loss, using a learning rate of $2e^{-6}$ for the pre-training of the model over English tweets, and $5e^{-7}$ for the fine-tuning over non-English tweets. We computed the performance on the development set after each training epoch, and kept the model obtaining the best performance. We used a batch size of 32. 
	
	We trained our models over 10 datasets and tested them over five different test sets in five languages. A summary of the datasets is shown in Table \ref{tab:datasets}. It was impossible to obtain the original datasets because of the nature of the data: if a tweet has been deleted, it should not be available online. Nevertheless, we think that our results are competitive since we are using state-of-the-art models and obtain better results than what are reported in the articles using the original datasets. For the English test set, which is the exact one used for SemEval-2017, our results are higher than the winner of the challenge. 
	
	For the French, German and Italian datasets, we used the same partition than the one used in the original challenges, with the tweets that were available online. For the Spanish dataset, the test sets were not available, hence we used the development set of TASS-2019 as test set, and the development set of TASS-2018 as development set. 
	For the Italian dataset, we selected the tweets from the general domain only and discarded the specific political tweets. 
	
	We computed metrics that are broadly employed for this kind of tasks in order to compare our models: the Average-Recall, the average of the F1 score between positive and negative example, as well as the macro F1 score 
	
	
	\begin{table}[]
		\centering
		\begin{tabular}{l|l|c|c|c|c}
			\textbf{Dataset}      & \textbf{Language}    & \textbf{Train} & \textbf{Dev} & \textbf{Test}  & \textbf{All}       \\ \hline \hline
			SB-10k  \cite{Cieliebak2017}     & German  &  4925    &  330   &    1315     &   6570       \\ \hline
			TASS-2019  \cite{Diaz-Galiano2019}  & \multirow{2}{*}{Spanish} &   \multirow{2}{*}{2133}   & \multirow{2}{*}{506}      &   \multirow{2}{*}{581}    &     \multirow{2}{*}{3220}  \\ 
			TASS-2018  \cite{Martinez-Camara2018}  &  &      &     &      &      \\ \hline
			DEFT-2015  \cite{Hamon2015}  & French  &   6489     &   407  &   2938  & 9427   \\ \hline			
			Sentipolc-16 \cite{Basile2016} & Italian &     6534  &   436  &     1964     &  8934  \\ \hline \hline
			SemEval-2017 \cite{Rosenthal2018} & \multirow{5}{*}{English} &   \multirow{5}{*}{47762}        &   \multirow{5}{*}{2000}   &  \multirow{5}{*}{12284}    & \multirow{5}{*}{62046}    \\ 
			SemEval-2013 \cite{Nakov2013} & &     &   & & \\
			SemEval-2014 \cite{Rosenthal2015a} &  &      &     &  &  \\
			SemEval-2015 \cite{Rosenthal2015b} &  &      &     &  &  \\ 
			SemEval-2016 \cite{Nakov2016} &  &      &     &   &   \\
		\end{tabular}
		\caption{Datasets used in our experiments}
		\label{tab:datasets}
	\end{table}
	
	
	\vspace*{-.1cm}
	
	\subsection{Results} \vspace*{-.1cm}
	
	
	The results of the experiments are shown in Table \ref{tab:results}. One important thing to note is that we do not compare our system to other state-of-the-art systems on those datasets. 
	This is due to the non availability of the complete datasets
	. The main contribution of this paper remains in the use of data-augmentation using automatic translation combined with a multilingual model pre-trained over English tweets. 
	
	Nevertheless, we believe that the results of the first configuration, without any pre-training neither data-augmentation are very competitive. For example, the best result reported by the authors of the SB10k has a $\text{F1}_\text{PN}$ of 65.09, which is below the performance of 67.1 we obtained with the Vanilla configuration. 
	
	The best results overall non-English languages are obtained using the pre-training as well as the data-augmentation technique. 
	
	%
	%
	%
	
	\begin{table}[]
		\centering
		\resizebox{.9\textwidth}{!}{%
			\begin{tabular}{l|c|c|c|ccc}
				Language          & Model &       Using English                 & D-A & $\text{Rec}_\text{avg}$ & $\text{F1}_\text{mac}$ & $\text{F1}_\text{PN}$ \\ \hline
				\multirow{5}{*}{English} & \cite{Cliche2017} (winner SemEval-2017)& \cmark                      & \xmark     &  68.1 &  $\varnothing$ &  68.5  \\
				& \cite{Nguyen2020} (SOTA)& \cmark                      & \xmark     &   \textbf{73.2}     &  $\varnothing$  & \textbf{72.8}     \\ \cline{2-7}
				
				& Monolingual & \cmark                      & \xmark     &   \textbf{72.8}     &  \textbf{71.7}  &  \textbf{72.3}    \\ \cline{2-7}
				& \multirow{2}{*}{Multilingual}   & \cmark    & \xmark     & 71.9   &  70.0 &   70.3  \\
				&    &       \cmark         &     \cmark         
				&       71.6     &   69.3   &   70.2  \\ \hline \hline 
				
				\multirow{3}{*}{German} & \multirow{3}{*}{Multilingual}   & \xmark                      & \xmark         &      72.6        &        73.9     &     67.1       \\
				&        &       \cmark      &    \xmark          &      74.1       &    \textbf{74.8} & \textbf{68.7}        \\
				&   &   \cmark         &     \cmark         &         \textbf{74.2}       &    74.7 & 68.5     \\ \hline 
				
				\multirow{3}{*}{Spanish} & \multirow{3}{*}{Multilingual}  & \xmark                      & \xmark         &    63.5          &      63.2       &       72.7     \\
				&       &          \cmark      &    \xmark          &     68.3        &      68.1  & 76.0      \\
				&      &      \cmark         &     \cmark         &      \textbf{69.8}      &    \textbf{69.6}    &    \textbf{78.2}         \\ \hline 
				
				\multirow{4}{*}{French} & Monolingual & \xmark                      & \xmark         &     72.9     &    72.8       &    71.6      \\ \cline{2-7}
				&  \multirow{3}{*}{Multilingual} & \xmark                      & \xmark         &      72.5        &    72.4        &    71.0        \\
				&    &   \cmark      &    \xmark          &    73.8          &  73.7      & 72.2      \\
				&     &    \cmark         &     \cmark         &    \textbf{74.4}         &     \textbf{74.5} & \textbf{72.8}       \\ \hline 	
				
				\multirow{3}{*}{Italian} & \multirow{3}{*}{Multilingual}  & \xmark                      & \xmark         &    63.0          &       60.7      &      55.3      \\
				&       &           \cmark      &    \xmark          &  67.1           &      64.4 &   60.2     \\
				&     &     \cmark         &     \cmark         &        \textbf{68.1}     &   \textbf{ 66.1}   & \textbf{62.0}     \\ \hline \hline 			
				\multirow{3}{*}{All (non English)} & \multirow{3}{*}{Multilingual}  & \xmark                      & \xmark         &         68.0    &     67.6        &     66.6      \\
				&       &           \cmark      &    \xmark          &     70.8      &   70.3   &    69.3  \\
				&     &     \cmark         &     \cmark         &  \textbf{71.6}     &  \textbf{71.2}    &   \textbf{70.4}  \\ \hline 		
			\end{tabular}%
		}
		\caption{Results of the different configuration. All the models were originally pre-trained over general text data.}
		\label{tab:results}
	\end{table} \vspace*{-.2cm}
	
	\vspace*{-1cm}
	\subsection{Analysis} 
	
	Because the original tweets datasets are not available online, it is difficult for us to compare with the results  in the literature for the datasets other than English. Nevertheless, the focus of our paper is not on beating the state-of-the-art but propose an method to use multilingual data to enhance the performance of a model using non-english data. Interestingly, we found better results than \cite{Nguyen2020} for both the RoBERTa and XLM-RoBERTa over SemEval-2017. We think that this may be the result of adjusting class weights in our loss function to manage imbalanced classes. 
	
	\paragraph*{Monolingual versus multilingual}
	
	As it is pointed out by \cite{Nguyen2020}, the best results over English are obtained using a monolingual model
	, when compared to the same multilingual model. Hence, RoBERTa reaches higher performances than its multilingual counterpart the XLM-RoBERTa. 
	This behavior is reproduced on the French datasets using CamemBERT, the French version of RoBERTa \cite{Martin2020}. Nevertheless, the pre-training of the multilingual model allow to obtain an increase in the performance of the French model. We think that this may be due to the lack of available examples in the target language. This confirms the hypothesis that pre-training a multilingual model with available data to use it on another language can be a good strategy to improve the results on a target language having less available examples for training. 
	
	\paragraph*{Effect of data-augmentation}
	
	Finally, the data-augmentation technique improve slightly the results for almost every language in different proportions. The biggest amelioration is obtained for Spanish, with an improvement of 1.5 points of the average recall compared with the model only pre-trained over English. The improvement over German is questionable. This may be due to the size of the dataset. The German train set is more than twice the size of the Spanish, which is the language were pre-training gives the better boost to the performances.

	\section{Conclusion} \vspace*{-.2cm}
	\label{sec:conclusion}
	
	We presented a technique that helps to improve the results of a sentiment analysis system over non-english tweets. We use multilingual model that is able to process external English data available in big quantities to pre-train the model, and machine translation to augment the dataset. This technique is simple and yet allows to take advantage of the multilingual models for non-English tweet datasets of limited size. 
	
	\newpage
	\bibliographystyle{coling}
	\bibliography{JRC.bib}

\end{document}